\documentclass{article}
\usepackage{spconf}
\usepackage{amsmath,graphicx}
\usepackage{booktabs}
\usepackage{amssymb}
\usepackage{multirow}
\title{AWDiff: An A trous Wavelet Diffusion Model for Lung Ultrasound Image Synthesis} 

\name{Maryam Heidari$^{1}$, Nantheera Anantrasirichai$^{1}$, Steven Walker$^{2}$, Rahul Bhatnagar$^{2}$ and Alin Achim$^{1}$}
\address{
    $^{1}$Visual Information Lab, University of Bristol, Bristol, UK \\
    $^{2}$Academic Respiratory Unit, Bristol Medical School, University of Bristol, Bristol, UK  
}

\begin{document}
\ninept
\maketitle

\begin{abstract}
Lung ultrasound (LUS) is a safe and portable imaging modality, but the scarcity of data limits the development of machine learning methods for image interpretation and disease monitoring. Existing generative augmentation methods, such as Generative Adversarial Networks (GANs) and diffusion models, often lose subtle diagnostic cues due to resolution reduction, particularly B-lines and pleural irregularities. We propose A trous Wavelet Diffusion (AWDiff), a diffusion-based augmentation framework that integrates the a trous wavelet transform to preserve fine-scale structures while avoiding destructive downsampling. In addition, semantic conditioning with BioMedCLIP, a vision–language foundation model trained on large-scale biomedical corpora, enforces alignment with clinically meaningful labels. On a LUS dataset, AWDiff achieved lower distortion and higher perceptual quality compared to existing methods, demonstrating both structural fidelity and clinical diversity.
\end{abstract}


\section{Introduction}

LUS has become an established modality for detecting pulmonary conditions such as pleural effusion, pneumonia, and pulmonary edema~\cite{Soldati2020,Yang2022Review}. Its portability, safety, and bedside accessibility make it particularly valuable in both emergency and routine clinical care. Nevertheless, the promise of machine learning for automated LUS interpretation remains under-realized, largely due to the limited size and heterogeneity of available datasets. Such constraints hinder the ability of algorithms to generalize across diverse patients, disease stages, and acquisition protocols.  
To address data scarcity, augmentation has been widely adopted. Conventional strategies such as geometric transformations and noise injection expand training sets but fail to reproduce the subtle imaging artifacts that are diagnostically critical in ultrasound.  Recent surveys emphasize a clear shift from simple geometric heuristics toward generative modeling as the diversity of data and deployment scenarios has expanded~\cite{Kazerouni2023Survey}.

Early efforts in ultrasound augmentation employed adversarial learning and autoencoding. Fatima \emph{et al.} proposed an autoencoder–GAN framework to address class imbalance while maintaining coarse anatomical structure~\cite{Fatima2025}. Conditional GANs have also been used at the lesion level to model targeted distribution shifts~\cite{Cai2024} and single-image approaches like SinGAN have similarly been explored, though with comparable limitations~\cite{Shaham2019SinGAN}. While these approaches demonstrated potential in small datasets, they generally provided limited semantic control and under-represented subtle, high-frequency, cues that carry diagnostic importance. Consequently, their impact on clinically robust pipelines has been modest. 

Diffusion models have more recently emerged as the state of the art for high-fidelity medical image synthesis. In musculoskeletal ultrasound, Balla \emph{et al.} reported improved segmentation when training with diffusion-generated cohorts compared to conventional augmentation~\cite{Balla2024}. Freiche \emph{et al.} fine-tuned latent diffusion models to generate realistic breast ultrasound images, validated through expert review~\cite{Freiche2025}. In echo-cardiography, Van De Vyver \emph{et al.} demonstrated that diffusion-driven augmentation enhanced segmentation robustness beyond standard baselines~\cite{Vyver2025}. Similar advantages have been observed in fetal imaging, where diffusion models improved fetal plane classification~\cite{Tian2025} and enabled anatomy-guided synthesis of normal and anomalous ultrasound images without abnormal exemplars~\cite{Duan2025}. 

More recently, diffusion models have also been applied to lung ultrasound, showing their potential for high-fidelity synthesis in this modality~\cite{Freiche2025SPIE}.
These studies collectively establish diffusion models as a reliable alternative to traditional augmentation techniques across diverse ultrasound modalities. Closer to lung ultrasound, Zhang \emph{et al.} adapted single and few-image diffusion (SinDDM/FewDDM) to expand limited cohorts, highlighting the potential of multi-scale denoising under severe data scarcity~\cite{Zhang2023LUS}. Beyond ultrasound, semantically conditioned diffusion has been applied in other modalities: Dorjsembe \emph{et al.} demonstrated controllable 3D MRI synthesis~\cite{Dorjsembe2024}, while Saragih \emph{et al.} generated paired data for endoscopic segmentation~\cite{Saragih2024}. Complementary physics-aware designs have also been explored, with Dominguez \emph{et al.} incorporating acoustic propagation priors to enhance realism in ultrasound generation~\cite{Dominguez2024PhysicsDiffusion}.

Overall, diffusion models have recently surpassed GANs in stability and fidelity~\cite{Wang2025,Vyver2025}, making them a state-of-the-art solution for medical image synthesis. However, some existing methods downsample during training or generation~\cite{Kulikov2023SinDDM}, thereby degrading the fine details that are indispensable for LUS interpretation, including pleural line continuity and the distribution of B-lines~\cite{Sakamoto2025,Shah2025}. Preserving such structures is essential if synthetic data are to hold clinical value.  

In this work, AWDiff directly tackles the two key limitations in prior approaches: (i) loss of fine structures from downsampling, avoided with a multi-scale a trous–wavelet encoder~\cite{Yu2016}; and (ii) weak semantic control, addressed through BioMedCLIP conditioning~\cite{Zhang2023} that aligns outputs with clinical attributes such as B-lines and pleural irregularities. Through this integration, AWDiff synthesizes images that are both anatomically faithful and pathologically diverse.

The remainder of this paper is structured as follows: section 2 details the AWDiff framework, including the forward and reverse diffusion processes. Section 3 describes the dataset and training setup with evaluations against existing methods. Finally, section 4 concludes with insights and directions for future work.

\begin{figure*}[!t]
\centering
\includegraphics[width=\linewidth]{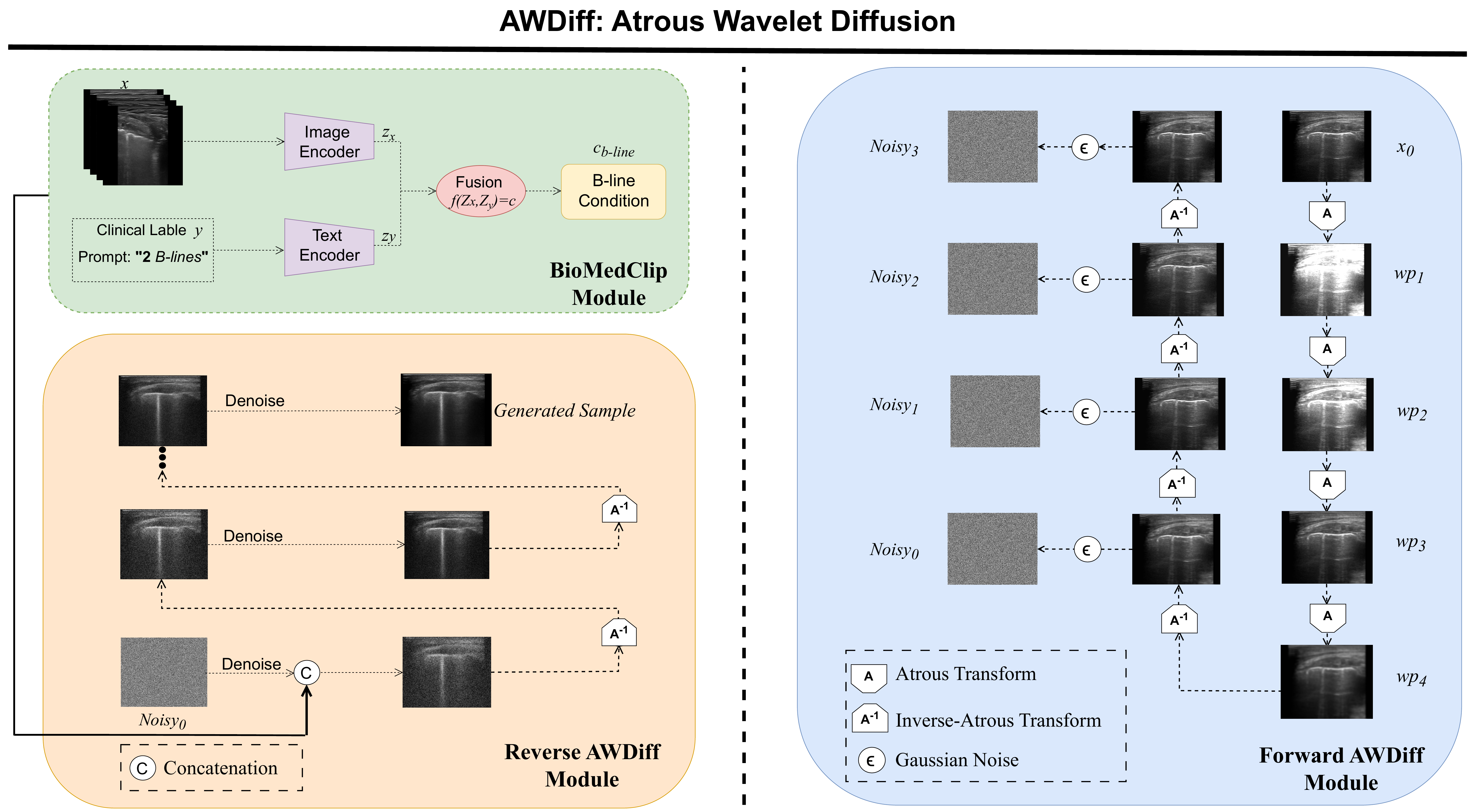}
\caption{Overview of the proposed AWDiff framework. In the forward diffusion stage, the input image is decomposed by a multi-scale a trous wavelet encoder into wavelet planes (wps) that preserve fine anatomical detail. These wavelet features are then fused with BioMedCLIP embeddings during the reverse diffusion process.}
\label{fig:method}
\end{figure*}

\section{AWDiff: Proposed Method}

AWDiff is a conditional diffusion framework for lung ultrasound images that preserves fine-scale structures while maintaining consistency with clinical labels. The model integrates multi-scale structural encoding with semantic guidance, enabling the generation of anatomically precise and diagnostically meaningful images. An overview of the framework is shown in Fig.~\ref{fig:method}.

\subsection{Problem Formulation}

Let $\mathcal{X}=\{x_i\}_{i=1}^N$ denote a dataset of LUS images, each associated with a clinical label $y_i \in \mathcal{Y}$ (e.g., B-line counts). The objective is to learn a conditional generative model:
\begin{equation}
p_\theta(x|y), \quad x \in \mathbb{R}^{H \times W},
\end{equation}

The reverse diffusion process is implemented using a UNet denoiser conditioned on semantic and structural information:
\begin{equation}
\epsilon_\theta(x_t,t,y,f)=
\text{UNet}\big(x_t,t,\text{CrossAttention}(z_y,f)\big),
\label{eq:cond_unet}
\end{equation}
where $z_y=\Phi_{\text{Text}}(y)$ denotes the BioMedCLIP text embedding and
$f=E_\psi(x)$ represents the multi-scale a trous wavelet features extracted from the input image.

AWDiff builds upon the denoising diffusion probabilistic model (DDPM)~\cite{ho2020denoising}. In the forward process, a clean image $x_0$ is gradually perturbed with Gaussian noise:
\begin{equation}
q(x_t|x_{t-1}) = \mathcal{N}\!\left(x_t;\sqrt{1-\beta_t}\,x_{t-1},\beta_t I\right),
\end{equation}
where $\{\beta_t\}_{t=1}^T$ defines the noise schedule. The process admits a closed-form marginal:
\begin{equation}
q(x_t|x_0) = \mathcal{N}\!\left(x_t;\sqrt{\bar{\alpha}_t}\,x_0,(1-\bar{\alpha}_t)I\right), 
\quad \bar{\alpha}_t=\prod_{s=1}^t(1-\beta_s).
\end{equation}

The reverse process is parameterized by a neural network:
\begin{equation}
p_\theta(x_{t-1}|x_t,y,f) =
\mathcal{N}\!\left(x_{t-1};\mu_\theta(x_t,t,y,f),\Sigma_\theta(x_t,t,y,f)\right),
\end{equation}
where $f$ denotes the multi-scale structural features extracted by the \`a trous–wavelet encoder (WP$_1$–WP$_S$ in Fig.~\ref{fig:method}). The denoiser $\epsilon_\theta$ is trained to predict the injected noise $\epsilon \sim \mathcal{N}(0,I)$ by minimizing the score-matching loss:
\begin{equation}
\mathcal{L}_{\text{MSE}} =
\mathbb{E}_{x_0,\epsilon,t}\,\big\|
\epsilon - \epsilon_\theta(x_t,t,y,f)\big\|^2.
\end{equation}

\subsection{A trous–Wavelet Encoder}

Preserving subtle diagnostic cues such as pleural continuity, vertical B-line artifacts, and speckle texture is critical for LUS augmentation. AWDiff employs an encoder $E_\psi$ that combines dilated convolutions with iterative wavelet decomposition.  

A trous convolution at layer $\ell$ with dilation rate $d$ is defined as:
\begin{equation}
\text{AtrousConv}(x)[i,j] =
\sum_{m,n} w^\ell_{m,n}\,x[i+d\cdot m, j+d\cdot n],
\end{equation}
which enlarges the receptive field while avoiding subsampling, thus preserving spatial resolution.  

Wavelet decomposition is then applied iteratively. Starting with $S^{(0)}=x$, smoothed images and wavelet planes are obtained as:
\begin{align}
S^{(s)} &= \text{AtrousConv}(S^{(s-1)},K_s), \\
WP^{(s)} &= S^{(s-1)} - S^{(s)},
\end{align}
where $K_s$ denotes the fixed B3-spline scaling filter dilated at scale $s$.

The final encoder representation is:
\begin{equation}
f = E_\psi(x) = \{WP^{(1)},WP^{(2)},\dots,WP^{(S)}\}.
\end{equation}
As shown in Fig.~\ref{fig:method}, these multi-scale wavelet planes capture fine structural detail and are injected into the denoiser at each reverse diffusion step to preserve clinically salient features.

\subsection{Semantic Conditioning with BioMedCLIP}

Structural preservation alone does not guarantee clinical reliability. AWDiff therefore employs semantic conditioning through BioMedCLIP~\cite{Zhang2023}, a vision–language model trained on large biomedical corpora.  

Given a label $y$, the text encoder produces an embedding $z_y=\Phi_{\text{Text}}(y)$, while the corresponding image encoder produces $z_x=\Phi_{\text{Image}}(x)$. The image embedding $z_x$ is used exclusively during training to compute a semantic alignment loss and does not participate in the conditioning of the reverse diffusion process. The reverse diffusion process is conditioned only on the text
embedding $z_y$ and the wavelet features $f$ (Eq.~\ref{eq:cond_unet}).
This mechanism (depicted in the fusion block of Fig.~\ref{fig:method}) ensures that the generated samples reflect both the structural fidelity of ultrasound and the semantic attributes encoded in clinical labels.

\subsection{Conditional Sampling}

Generation starts from Gaussian noise $x_T \sim \mathcal{N}(0,I)$ and proceeds by iterative denoising:
\begin{equation}
x_{t-1} = \mu_\theta(x_t,t,z_y,f) + \Sigma_\theta^{1/2}\epsilon,
\quad \epsilon \sim \mathcal{N}(0,I).
\end{equation}
As indicated by the reverse arrows in Fig.~\ref{fig:method}, multi-scale wavelet features are injected throughout the trajectory, preserving high-frequency details that are typically lost in conventional diffusion sampling. This design ensures that pleural line morphology and B-line artifacts remain distinct and clinically plausible.

\subsection{Loss Functions}

The training objective balances structural fidelity with semantic alignment. The primary supervision is the denoising score-matching loss:
\begin{equation}
\mathcal{L}_{\text{MSE}} =
\mathbb{E}_{x_0,\epsilon,t}\,\big\|
\epsilon - \epsilon_\theta(x_t,t,y,f)\big\|^2.
\end{equation}

To further enforce semantic consistency, it adds a BioMedCLIP-based loss:
\begin{equation}
\mathcal{L}_{\text{BioMedCLIP}} =
1 - \cos\!\big(\Phi_{\text{Image}}(\hat{x}_0(t)), \Phi_{\text{Text}}(y)\big), 
\end{equation}
where $\hat{x}_0$ denotes the reconstructed sample at step $t$.  

The final objective is a weighted combination:
\begin{equation}
\mathcal{L}_{\text{total}} =
\mathcal{L}_{\text{MSE}} + \lambda_1 \mathcal{L}_{\text{BioMedCLIP}}.
\end{equation}

This formulation enforces both pixel-level accuracy and semantic alignment, mitigating the mode collapse issues often observed in GAN-based methods and ensuring that synthetic LUS images remain diagnostically reliable.

\section{Experimental Results}

\subsection{Dataset and Training Setup}
Our evaluation was conducted on a clinical dataset of 360 dialysis-related LUS scans~\cite{Yang2023BLine,Anantrasirichai2017}, each paired with expert-annotated labels. To expand this limited corpus, AWDiff was used to synthesize additional samples, yielding a total of 2,260 images for fine-tuning BioMedCLIP. All images were resized to $256\times256$ pixels for consistency. Clinical labels were converted into text prompts for conditioning, while BioMedCLIP weights were frozen to avoid catastrophic forgetting.  
Training employed the Adam optimizer ($\alpha=10^{-4}, \beta_1=0.9, \beta_2=0.999$) with a linear $\beta_t$ schedule over $T=100$ diffusion steps. Models were trained for 120k epochs with a batch size of 128. To enhance stability, an exponential moving average (EMA) of model parameters was maintained during sampling. This setup ensured that results reflected the generative capacity of AWDiff rather than fluctuations in optimization.

\subsection{Evaluation and Analysis}
Performance was evaluated using three complementary metrics: SIFID~\cite{Kulikov2023SinDDM}, LPIPS~\cite{zhang2018lpips}, and NIMA~\cite{talebi2018nima}. SIFID evaluates structural fidelity through the alignment of feature distributions from real and generated images, while LPIPS captures perceptual similarity using deep representations. To complement these measures, NIMA extends the evaluation to aesthetic quality, reflecting predicted human ratings of visual appeal.
\begin{table}
\centering
\caption{Comparison of SIFID, LPIPS, and NIMA for SinDDM, SinGAN, and AWDiff.}
\label{tab:sinddm_steps}
\begin{tabular}{lcccc}
\toprule
\textbf{Method} & \textbf{Steps} & \textbf{SIFID}$\downarrow$ & \textbf{LPIPS}$\uparrow$ & \textbf{NIMA}$\uparrow$ \\
\midrule
\multirow{2}{*}{SinDDM} 
 & 80k  & 0.05$\pm$0.08 & 0.26$\pm$0.04 & 5.12$\pm$0.5 \\
 & 120k & 0.04$\pm$0.01 & 0.27$\pm$0.02 & 5.38$\pm$0.1 \\
\midrule
SinGAN 
 & Final (16k) & 0.08$\pm$0.01 & 0.21$\pm$0.06 & 4.3$\pm$0.1 \\
\midrule
\multirow{2}{*}{AWDiff} 
 & 80k  & 0.03$\pm$0.5 & 0.36$\pm$0.09 & 5.24$\pm$0.6 \\
 & 120k & \textbf{0.03$\pm$0.06} & \textbf{0.37$\pm$0.02} & \textbf{5.45$\pm$0.7} \\
\bottomrule
\end{tabular}
\end{table}
\begin{figure*}[!tbp]
\centering
\includegraphics[width=\textwidth]{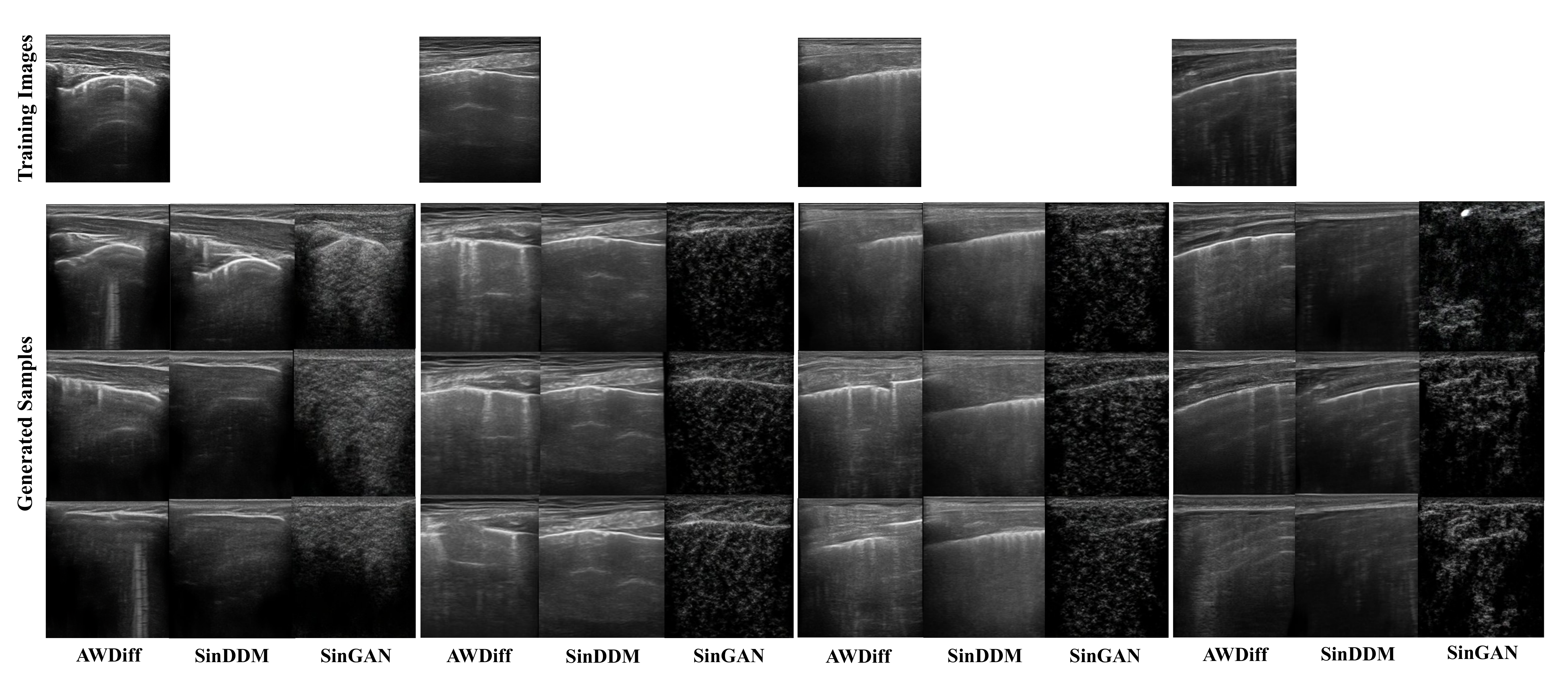}
\caption{Comparison of AWDiff, SinDDM, and SinGAN on four training images, with three generated samples per method. AWDiff more faithfully preserves B-line structures compared to SinDDM and SinGAN.}
\label{fig:qualitative}
\end{figure*}
According to Table~\ref{tab:sinddm_steps}, AWDiff achieves the best overall performance at 120k steps (SIFID = 0.03, LPIPS = 0.37, NIMA = 5.45), outperforming both SinDDM and SinGAN. Compared with SinDDM, AWDiff shows improved metrics, with much of its advantage already apparent at 80k. By contrast, SinGAN performs markedly worse, highlighting its limited suitability for this medical imaging task. Noting that 16k represents SinGAN’s terminal training steps, beyond which further iterations lead to overfitting.

Also, representative qualitative results are shown in Fig.~\ref{fig:qualitative}. For four input images, it displays three samples per method. SinDDM and SinGAN often produce weakened pleural or B-lines, whereas AWDiff better preserves pleural continuity and sharper vertical features. Qualitative feedback from clinical experts suggested that AWDiff outputs are visually easier to interpret.

\begin{figure}[!htbp]
\centering
\includegraphics[width=\linewidth]{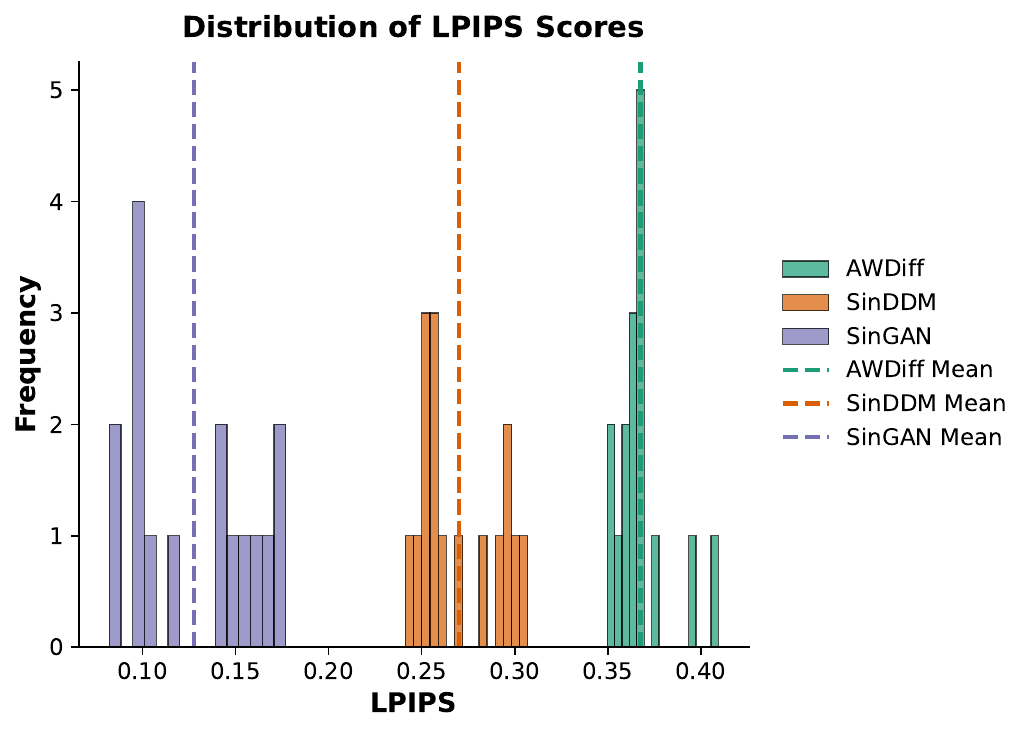}
\caption{LPIPS score for AWDiff, SinDDM, and SinGAN.}
\label{fig:lpips_hist}
\end{figure}

Perceptual diversity was further examined via LPIPS distributions over 16 samples (Fig.~\ref{fig:lpips_hist}). AWDiff yields a more favorable distribution than SinDDM and SinGAN which indicates both higher fidelity to the reference images and controlled variability across generated samples.

\medskip
Since AWDiff explicitly leverages a wavelet encoder, we further compared its structural preservation against standard discrete wavelet transform (DWT)~\cite{Alessio2015DSP}, a widely used standard tool in multiresolution analysis.
Accordingly, we employed the Complex Wavelet Structural Similarity (CW-SSIM)~\cite{Wang2005CWSSIM}, a metric specifically designed to evaluate local structural similarity in the wavelet domain. 
This choice is motivated by the fact that clinically relevant patterns in LUS, such as vertical B-lines, are strongly characterized by directional wavelet coefficients. These coefficients emphasize localized orientation and scale information, which are critical for capturing the sharp artifacts that underpin diagnostic interpretation. 

\begin{figure}[!htbp]
\centering
\includegraphics[width=\linewidth]{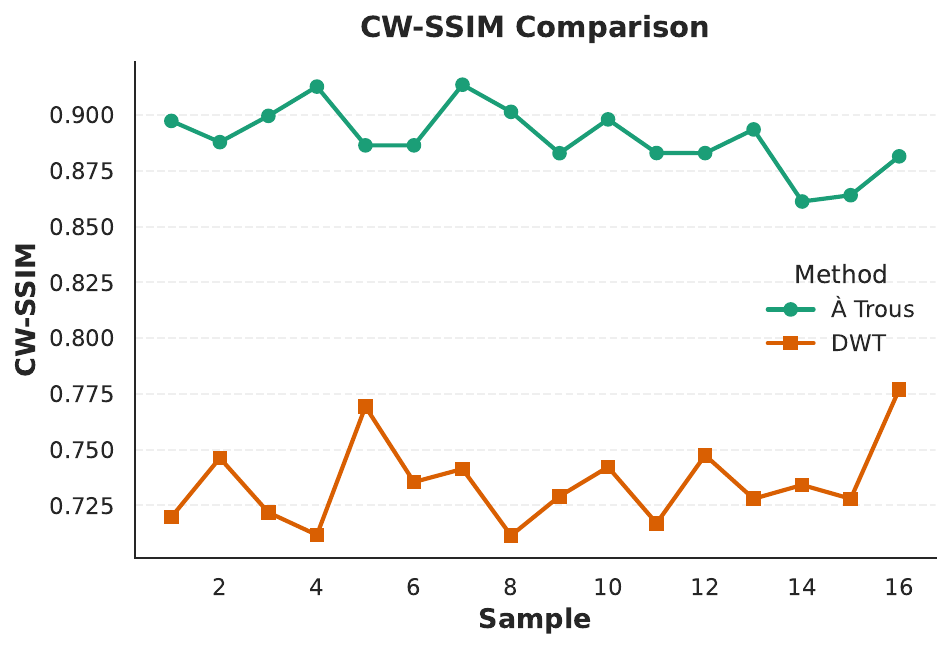}
\caption{CW-SSIM comparison between \`a trous and DWT.}
\label{fig:cwssim}
\end{figure}

Fig.~\ref{fig:cwssim} presents the CW-SSIM distributions for 16 generated samples. Across all cases, the a trous wavelet consistently yields higher similarity scores than DWT. This advantage reflects a stronger capacity to retain local structures and subtle textural cues, which are particularly critical in LUS imagery. 

\section{Conclusion}
This work introduces AWDiff, a wavelet-conditioned diffusion framework for lung ultrasound synthesis. Combining a multi-scale a trous wavelet encoder with BioMedCLIP semantic conditioning, AWDiff preserves fine diagnostic structures while aligning outputs with clinically meaningful labels. Compared with SinDDM and SinGAN, it consistently attains lower SIFID, higher LPIPS, and superior NIMA scores, with experts noting sharper and more reliable B-line reproduction. Beyond surpassing prior methods, AWDiff offers a principled route to generating diverse yet structurally precise synthetic cohorts, directly addressing the scarcity of LUS data. These results position wavelet-driven diffusion as a powerful tool for medical image synthesis, with potential to improve downstream diagnostic models. 

%\bibliography{refs}
\bibliographystyle{IEEEbib}

\end{document}